\pdfoutput=1
\documentclass[11pt]{article}

\usepackage[preprint]{acl}

\usepackage{times}
\usepackage{latexsym}

\usepackage[T1]{fontenc}
\usepackage[utf8]{inputenc}

\usepackage{microtype}

\usepackage{inconsolata}

\usepackage{graphicx}

\usepackage{tabularx}
\usepackage{booktabs}
\usepackage{multirow}
\usepackage{amsmath}
\usepackage{amssymb}
\usepackage{makecell}

\title{Multi-Aspect Knowledge Distillation for Language Model with \\ Low-rank Factorization}

\author{Zihe Liu$^1$$^2$$^*$, Yulong Mao$^1$$^2$\thanks{Equal Contribution}, Jinan Xu$^1$$^2$, Xinrui Peng$^2$, Kaiyu Huang$^1$$^2$ \\  
  $^1$ Key Laboratory of Big Data \& Artificial Intelligence in Transportation\\
  $^2$ School of Computer Science and Technology, Beijing Jiaotong University \\
}

\begin{document}
\maketitle
\begin{abstract}
% 知识蒸馏是一种有效的预训练语言模型压缩技术。
% 然而，现有方法仅关注层间的知识分布，这可能会在对齐过程中造成的更细粒度信息的缺失。
% 为了解决这个问题，我们引入了多层次知识蒸馏（MAKD）方法。MAKD 更深入地模拟自关注和前馈模块，捕获不同层次丰富的语言知识信息。
% 实验结果表明，在相同存储参数预算的情况下，与各种强基线相比，MAD可以获得具有竞争力的性能。
% 此外，我们的方法在对自回归架构模型的蒸馏上同样有着出色的表现。
Knowledge distillation is an effective technique for pre-trained language model compression. However, existing methods only focus on the knowledge distribution among layers, which may cause the loss of fine-grained information in the alignment process.
To address this issue, we introduce the Multi-aspect Knowledge Distillation (MaKD) method, which mimics the self-attention and feed-forward modules in greater depth to capture rich language knowledge information at different aspects. 
Experimental results demonstrate that MaKD can achieve competitive performance compared with various strong baselines with the same storage parameter budget.
In addition, our method also performs well in distilling auto-regressive architecture models.
We have made our code and model publicly accessible at [https://anonymous].
\end{abstract}

\section{Introduction}
% 基于变压器的预训练语言模型已经在各种自然语言处理（NLP）任务中取得了显著的成功。
% 它们首先在大规模的无监督文本语料库上进行训练，然后对下游任务进行微调。
Transformer-based Pre-trained Language Models~(PLMs)~ \cite{devlin-etal-2019-bert, lewis-etal-2020-bart, liu2019robertarobustlyoptimizedbert, NEURIPS2019_dc6a7e65, he2021debertadecodingenhancedbertdisentangled} have achieved remarkable performance across a wide range of Natural Language Processing~(NLP) tasks. These models are initially trained on large-scale unsupervised text corpora and subsequently fine-tuned for specific downstream applications \cite{Lee_2019, mars2022word, raffel2023exploringlimitstransferlearning}.
% 然而，plm在存储、内存和计算时间方面的成本很高，这给现实应用中的在线服务带来了挑战。因此，在保持plm性能的同时对其进行压缩是至关重要和可行的。
However, PLMs often incur high costs in terms of storage, memory, and computation time, posing substantial challenges for their deployment in real-world and online services. Therefore, it is crucial and feasible to to explore techniques for compressing PLMs while preserving their performance \cite{li2020trainlargecompressrethinking}.

\begin{figure}[t]
    \centering
    \includegraphics[width=\linewidth]{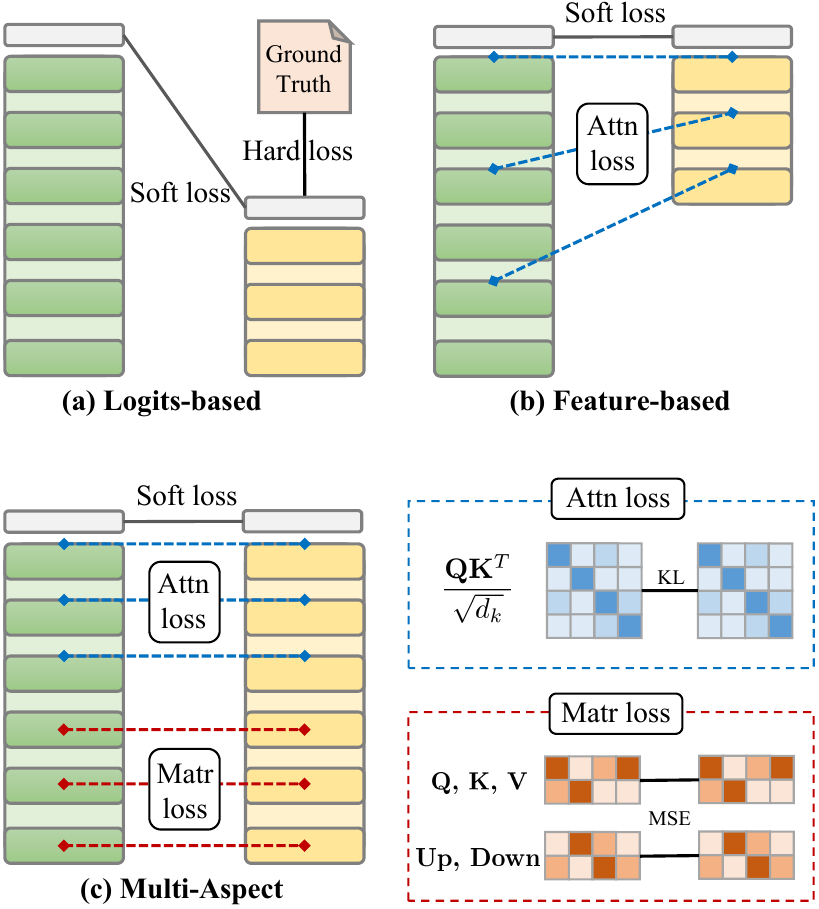}
    \caption{A comparison of logits-based (a), feature-based (b) and our multi-aspect (c) learning method.}
    \label{fig:intro}
\end{figure}

% 先提出Matrix Distillation，然后为了保证层间对应没有映射gap，才会用SVD分解初始化模型。
Knowledge distillation~(KD)~\cite{hinton2015distillingknowledgeneuralnetwork, romero2015fitnetshintsdeepnets} has become a widely adopted technique for compressing pre-trained transformers, transferring knowledge from a large~(teacher) model to a smaller~(student) model by minimizing the discrepancy between their feature representations. 
% 在以往的压缩方法中，KD有两个主流方面：
As shown in Figure~\ref{fig:intro}, previous KD methods primarily focus on two main approaches:
(1) \textbf{Logits-based} measures the divergence of the soft target probabilities between student and teacher models \cite{hinton2015distillingknowledgeneuralnetwork, tang2019distillingtaskspecificknowledgebert, sanh2020distilbertdistilledversionbert, turc2019wellreadstudentslearnbetter, gu2024minillm};
(2) \textbf{Feature-based} aims to align the intermediate features including token embeddings, hidden states, and self-attention distributions \cite{sun-etal-2019-patient, sun-etal-2020-mobilebert, jiao-etal-2020-tinybert, NEURIPS2020_3f5ee243, tahaei-etal-2022-kroneckerbert}.
% 然而，中间层方面的知识信息粒度过粗，且不利于表征多方面（multi-aspect）知识。
However, the knowledge extracted from intermediate layers tends to be coarse-grained, where the attention weight is the interaction result of Q and K, while aligning the matrices (e.g., Q/K/V) separately can retain the lower-level feature expressions.
% 模仿中间层可能丢失部分信息，因为注意力权重是 Q 和 K 的交互结果，而单独对齐 Q/K/V 能保留更底层的特征表达。
% \lzh{However, imitating the intermediate layers may lose some information because the attention weight is the interaction result of Q and K, while aligning Q/K/V separately can retain the lower-level feature expressions.}
% 与此同时，一些研究表明，预训练语言模型的self-attention maxtrix 捕获了丰富的语言信息层次结构，并且浅层和深层网络蕴含的信息粒度不同（Jawahar 2019）。
Meanwhile, recent studies have demonstrated that the self-attention matrix in pre-trained language models captures a rich, hierarchical structure of linguistic information, with varying granularities across shallow and deep layers \cite{jawahar-etal-2019-bert, clark-etal-2019-bert, yan-etal-2024-effective}.
% 因此，这使我们产生了一个直观的想法：我们能否通过层内矩阵方面的细粒度和层间信息相结合来帮助对齐教师和学生的分布？
As a result, this observation leads to an intuitive question: \textit{Can we combine the fine-grained knowledge from the intra-layer self-attention matrix with intermediate layer information to improve the alignment of the distributions between the teacher and the student models?}

% 在这项工作中，我们提出了多方面知识蒸馏（MAKD）， which 将层内矩阵中的细粒度知识与中间层信息相结合。
In this work, we propose \textbf{M}ulti-\textbf{a}spect \textbf{K}nowledge \textbf{D}istillation~(MaKD), a novel approach which combines fine-grained knowledge from the intra-layer self-attention matrix with intermediate layer information.
% 我们引入了细粒度矩阵方面的蒸馏，它深度模仿了自关注模块和前馈模块。
% We introduce a fine-grained a fine-grained matrix aspect distillation, which mimics the self-attention and feed-forward modules in greater depth.
% 为了实现层内蒸馏，我们需要消除教师模型和学生模型在不同transformers层之间投影时产生的gap。我们follow BERT-SVD、FWSVD、KroneckerBERT等研究工作，通过对教师模型进行SVD低秩矩阵分解来获得初始化的学生模型，这同时也能避免删除整个层可能会导致模型中包含的重要信息被丢弃。
To achieve effective intra-layer distillation, we first eliminate the gap that occurs when projecting across different Transformer layers between the teacher and the student models. 
% We initialize the student model by performing low-rank matrix factorization on the teacher model. This approach ensures that important information is not discarded, as might occur if entire layers are removed.
% 
In particular, we initialize the student model through Singular Value Decomposition (SVD) of matrices, which reduces the number of parameters while preserving critical information from each layer. 
Subsequently, during the task-agnostic pre-training phase, we conduct distillation at three different granularity aspects using a small amount of corpus. In the Matrix Distillation stage, we align the linear mapping layers in the Multi-Head Attention~(MHA) and Feed-Forward Neural Network~(FNN) modules. In the Layer Distillation stage, we capture the representational differences between self-attention distributions. 
In the Model Distillation stage, we minimize the soft cross-entropy loss of the logits between the teacher and student models.
% Firstly, we initialize the student model through Singular Value Decomposition (SVD) of matrices. This enables the reduction of the number of parameters while retaining the information in each layer. Subsequently, during the task-agnostic pre-training phase, we conduct distillation at three different granularity levels using a small amount of corpus. In the Matrix Distillation stage, we align the linear mapping layers in the Multi - Head Attention (MHA) and Feed-Forward Neural Network (FNN) modules. In the Layer Distillation stage, we capture the representational differences between self-attention distributions. In the Model Distillation stage, we penalize the soft cross - entropy loss of the logits between the teacher and student networks.
% 实验
Experimental results demonstrate that the MaKD outperforms various baselines for different student model sizes. Specifically, the 12-layer model with 768 hidden dimensions distilled from BERT$\rm_{\text{BASE}}$ is 2.0× faster, while maintaining more than 99\% average accuracy on SQuAD and several GLUE benchmark tasks. 
Moreover, the auto-regressive model distilled from GPT-2$\rm_{\text{BASE}}$ and LLaMA-3 also achieves competitive performance on instruction-following tasks, with significantly fewer Transformer parameters.

To sum up, our contributions are as follows:

\begin{itemize}
    \item We introduce a novel distillation framework, Multi-aspect Knowledge Distillation (MaKD), which achieves competitive performance through a hierarchical distillation strategy. 
    \item We propose fine-grained matrix distillation which improves knowledge alignment during the distillation process by capturing detailed aspects of the internal representations of the model.
    \item Our approach is applicable to various Transformer-based models, including both encoder-only and auto-regressive decoder-only architectures with different sizes.
\end{itemize}

\begin{figure*}[t]
    \centering
    \includegraphics[width=\linewidth]{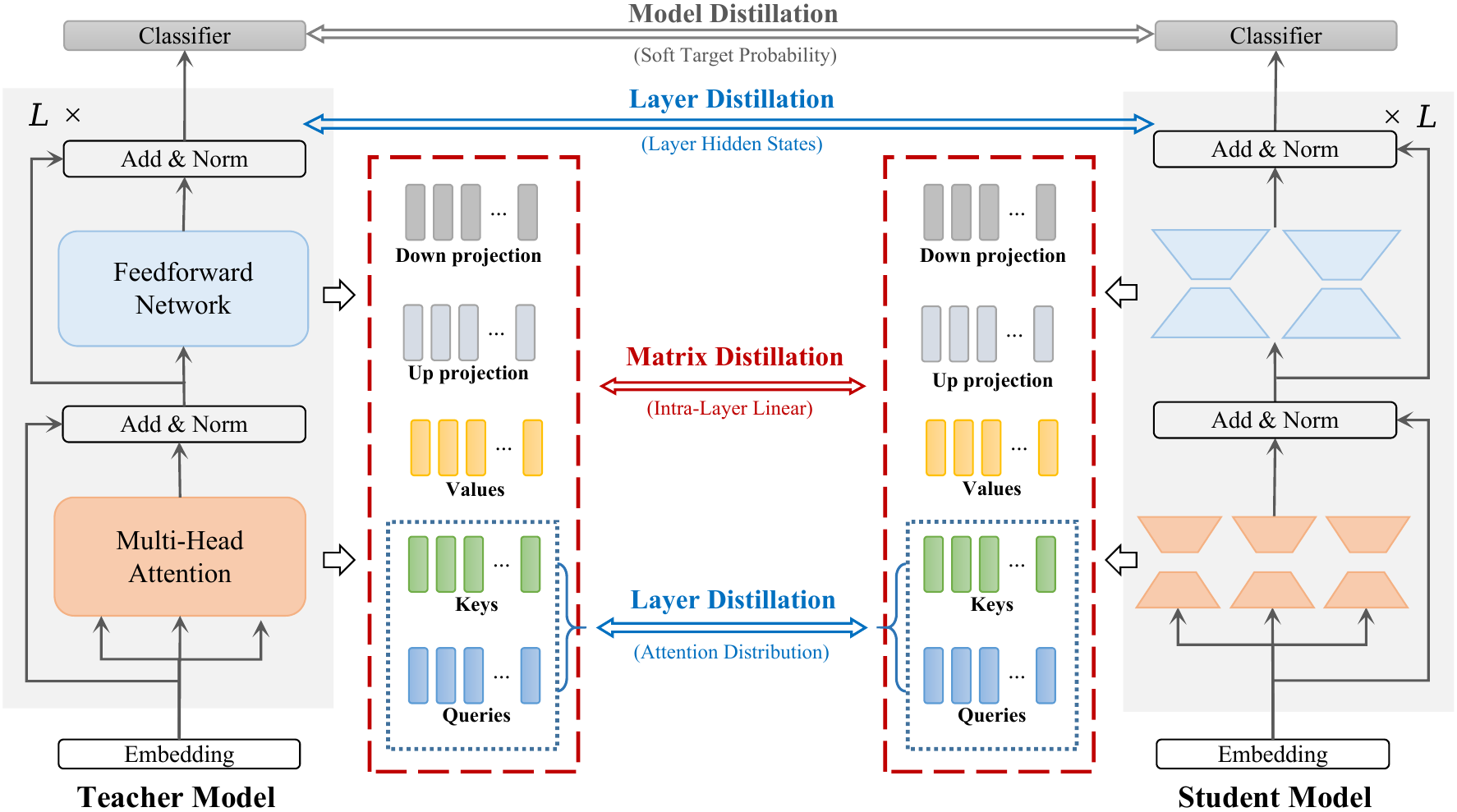}
    \caption{Overview of multi-aspect knowledge distillation. We obtain student model with the same hidden dimensions as the teacher model by low-rank matrix decomposition. We introduce matrix distillation computed by intra-layer linear, which consists of MHA mappings (queries, keys and values) and FNN vectors (up projection and down projection). We choose multi-aspect hierarchical distillation to achieve a balance between performance and training speed.}
    \label{fig:model}
\end{figure*}
% 我们通过低秩矩阵分解获得隐藏维度与教师模型相同的学生模型
% 我们选择多层面蒸馏来实现性能与训练速度之间的平衡

\section{Backgrounds}
\subsection{Transformer Layer}
Multi-layer Transformer \cite{NIPS2017_3f5ee243} has been widely adopted in pre-trained models. 
The input vector $(\{\mathbf{x}_{i}\}_{i = 1}^{\vert x\vert})$ are packed to $\mathbf{H}^{0} = [\mathbf{x}_{1},\cdots,\mathbf{x}_{\vert x\vert}]$. After that, the $L$-layer transformer computes the encoding vectors following:
\begin{equation}
    \mathbf{H}^{l} = \text{Transformer}_{l}(\mathbf{H}^{l - 1}),\ l \in [1, L]
\end{equation}
The final output $\mathbf{H}^{L} = [h_{1}^{L},\cdots,h_{|x|}^{L}] \in \mathbb{R}^{|x|\times d}$ is employed as the contextualized representation of $\{\mathbf{x}_{i}\}_{i = 1}^{\vert x\vert}$. 
Each transformer layer consists of a multi-head self-attention (MHA) sub-layer and a feed-forward (FNN) sub-layer. In these two sub-layers, the residual connection \cite{7780459} is employed, followed by Layer Normalization (LN) \cite{ba2016layernormalization}.

\paragraph{MHA} For the $l$-th transformer layer with $A_{h}$, the output $\mathbf{O}_{l,a}$ of the self-attention head  $a\in[1, A_{h}]$ is calculated via:
\begin{equation}
    \begin{split}
    \mathbf{Q}_{l,a} = \mathbf{H}^{l - 1}\mathbf{W}_{l,a}^{Q}\\
    \mathbf{K}_{l,a} = \mathbf{H}^{l - 1}\mathbf{W}_{l,a}^{K}\\
    \mathbf{V}_{l,a} = \mathbf{H}^{l - 1}\mathbf{W}_{l,a}^{V}
    \end{split}
\end{equation}
\begin{equation}
    \mathbf{A}_{l,a}={\rm softmax}(\dfrac{\mathbf{Q}_{l,a}\mathbf{K}_{l,a}^{T}}{\sqrt{d_{k}}})
\end{equation}
\begin{equation}
    \mathbf{O}_{l,a}=\mathbf{A}_{l,a}\mathbf{V}_{l,a}
\end{equation}
Previous layer's output $\mathbf{H}^{l-1}\in \mathbb{R}^{|x|\times d}$ is linearly projected to a triple of queries, keys and values using parameter matrices $\mathbf{W}^{Q}_{l,a}, \mathbf{W}^{K}_{l,a}, \mathbf{W}^{V}_{l,a}\in \mathbb{R}^{d\times d_{k}}$, respectively.
$|x|$ represents the length of input sequence. $A_{h}$ and $d$ indicate the number of attention heads and hidden size. $d_{k}=\frac{d}{A_h}$ is the dimension of each head. The final output of MHA sub-layer is as follows:
\begin{equation}
    \mathbf{O}_l = \text{LN}(\mathbf{H}^{l - 1} + (\lVert_{a = 1}^{A}\mathbf{O}_{l,a})\mathbf{W}_l^O)
\end{equation}
where $\mathbf{W}_l^O\in \mathbb{R}^{d\times d}$, $\text{LN}$ is layer normalization and $\lVert$ denotes the concatenation operation.

\paragraph{FNN}
The $l$-th FFN sub-layer consists of an up projection and a down projection, parameterized by $\mathbf{W}^{U}_{l}\in \mathbb{R}^{d\times d_{f}}$, $\mathbf{W}^{D}_{l}\in \mathbb{R}^{d_{f}\times d}$, and corresponding bias $\mathbf{b}^{U}_{l}\in \mathbb{R}^{d_{f}}$, $\mathbf{b}^{D}_{l}\in \mathbb{R}^{d}$. 
\begin{equation}
    \text{FFN}(\mathbf{O}_{l}) = \text{gelu}(\mathbf{O}_{l}\mathbf{W}_{l}^{U} + \mathbf{b}_{l}^{u})\mathbf{W}_{l}^{D} + \mathbf{b}_{l}^{d}
\end{equation}
Tipically, $d_{f}=4d$. Finally, we obtain the output of layer $l$ by:
\begin{equation}
    \mathbf{H}^{l} = \text{LN}(\mathbf{O}_{l} + \text{FFN}(\mathbf{O}_{l}))
\end{equation}

\subsection{Knowledge Distillation}
Knowledge Distillation (KD) trains a compact student model $S$ by mimicking the behaviors of the teacher model $T$. The losses can be categorized into logit-based and feature-based.

For logit-based loss, the target is to minimize the distance between logit distribution $\mathbf{p}_s$ from the student and $\mathbf{p}_t$ from the teacher, which can be formalized as:
\begin{equation}
    \mathcal{L}_{logit} = \mathcal{H}_1(\mathbf{p}_s/\tau, \mathbf{p}_t/\tau)
\end{equation}
where $\tau$ is the temperature and $\mathcal{H}_1$ is the cross-entropy loss or KL-divergence.

Feature-based loss aims to align the intermediate features between the teacher and the student by:
\begin{equation}
    \mathcal{L}_{feature} = \mathcal{H}_2(f^S(x), f^T(x)),
\end{equation}
where $\mathcal{H}_1$ is the loss function such as Mean Square Error (MSE) and $f(x)$ denotes for the intermediate output including hidden state vector and attention distribution.

\section{Multi-Aspect Distillation}
Figure \ref{fig:model} gives an overview of the multi-aspect knowledge distillation.
In this section, we propose a novel distillation method for Transformer-based models, and present a multi-aspect learning framework for our model distilled, which is called MaKD.

\subsection{Low-Rank Factorization}
Transformer mainly consist of two components: the Multi-Head Attention (MHA) and the Feed-Forward Neural Network (FNN). We first use singular value decomposition (SVD) to obtain an equivalent form of $\mathbf{W}\in \mathbb{R}^{n\times m}$ as the product of three matrices:
\begin{equation}
    \mathbf{W}=\mathbf{U}\mathbf{\Sigma} \mathbf{V}^{\rm T}
\end{equation} 
where $\mathbf{U}\in \mathbb{R}^{n\times r}$, $\mathbf{\Sigma} \in \mathbb{R}^{r\times r}$, $\mathbf{V}\in \mathbb{R}^{r\times m}$ and $r$ is the rank of matrix $\mathbf{W}$. $\mathbf{\Sigma}$ is a diagonal matrix of nonzero singular values $\{\sigma_{1}, \sigma_{2}, ..., \sigma_{r}\}$ in descending order. Then, low-rank approximation with targeted rank $k$ is obtained by keeping the top-k singular values in $\mathbf{\Sigma}$ as well as their corresponding column vectors in $\mathbf{U}$ and $\mathbf{V}$ :
\begin{equation}
    \mathbf{W}\approx \mathbf{U}_{[:,:k]}\mathbf{\Sigma}_{[:k,:k]}\mathbf{V}^{T}_{[:,:k]} = \mathbf{A}\mathbf{B}
\end{equation}
where $\mathbf{A} = \mathbf{U}_{[:,:k]}\mathbf{\Sigma}_{[:k,:k]}$ and $\mathbf{B} = \mathbf{V}^{T}_{[:,:k]}$ are the two final sub-matrices of which the product is used to replace $\mathbf{W}$. After such low-rank factorization, the number of parameters is reduced from $nm$ to $k(n + m)$. Different compression rates can be achieved by varying the preserved rank $k$.

\subsection{Transformer Distillation}
\paragraph{Layer Distillation}
% 一些工作已经证明了self-attention可以捕获丰富的语义信息。
% 与之前的工作通常将教师和学生之间不同层的中间特征进行对齐不同，我们通过上述的svd分解消除了在不同层间映射时产生的gap。
% 具体来说，我们逐层最小化教师和学生之间的自注意力分布的KL散度
Some works have show that self-attention distributions of pre-trained LMs capture a rich hierarchy of linguistic information \cite{jawahar-etal-2019-bert, clark-etal-2019-bert}. Different from previous works which usually align the intermediate features of different layers between the teacher and the student, we have eliminated the gap arising from mapping across different layers through the above-mentioned matrix factorization. Specifically, we minimize the KL-divergence between the self-attention distributions of the teacher and student layer-by-layer:
\begin{equation}
    \mathcal{L}_{\rm attn}=\frac{1}{A_{h}}\sum_{a=1}^{h}D_{KL}(\mathbf{A}^{S}_{L,a}|| \mathbf{A}^{T}_{L,a})
\end{equation}
where $A_h$ is the number of attention heads, $\mathbf{A}_{i}\in \mathbb{R}^{|x|\times |x|}$ refers to the attention matrix corresponding to the $i$-th head of teacher or student, $|x|$ is the input text length. 

% 除了基于注意力的提炼，我们还从Transformer层的输出中提炼知识，目标如下：
In addition to the attention based distillation, we also distill the knowledge from the output of Transformer layer, and the objective is as follows:
\begin{equation}
    \mathcal{L}_{\rm hidn}=MSE(\mathbf{H}^{S}_{L}, \mathbf{H}^{T}_{L})
\end{equation}
where the matrices $\mathbf{H}^{S}\in \mathbb{R}^{|x|\times d} $ and  $\mathbf{H}^{T}\in \mathbb{R}^{|x|\times d}$  refer to the hidden states of student and teacher networks respectively, which are calculated by Equation 7. $d$ denote the hidden size of model.

The layer aspect training loss is computed via summing the attention loss and hidden loss:
\begin{equation}
 \mathcal{L}_{\rm layer}=\mathcal{L}_{\rm attn}+\mathcal{L}_{\rm hidn}
\end{equation}

\paragraph{Matrix Distillation}
% 除了每一层之间的输出，我们提出使用层内的矩阵信息去指导学生的训练。
% 层内的矩阵信息通过
In addition to the outputs between each layer, we propose to use the in-layer matrix information in self-attention and fully connected module to guide the training of the student model.
Specifically, the multi-head self - attention (MHA) mechanism plays a crucial role, which enables the model to capture different aspects of the input sequence simultaneously.
The loss between the multi-head self-attention matrix of the teacher and student is used as the training objective:
\begin{equation}
    \mathcal{L}_{\rm MHA} =\frac{1}{A_{h}}\sum_{a=1}^{A_{h}}MSE(\mathbf{M}^{S}_{L,a}, \mathbf{M}^{T}_{L,a})
\end{equation}
where $\mathbf{M}^{S}_{L,a}\in \mathbb{R}^{|x|\times d}$ and $\mathbf{M}^{T}_{L,a}\in \mathbb{R}^{|x|\times d}$ are the queries, keys and values of an attention head in self-attention module for the student's and teacher's Transformer layer. $A_{h}$ denotes the number of attention heads, which determines the model's ability to capture diverse information from different perspectives. $d$ is the hidden dimension of model.

The feed-forward neural network (FNN) module in the Transformer architecture further processes the output of the MHA module through a series of linear transformations. We calculate the loss between the FNN modules of the teacher and student models in a similar way. The loss formula is:
\begin{equation}
    \mathcal{L}_{\rm FNN} = \frac{1}{A_{h}}\sum_{a=1}^{A_{h}}MSE(\mathbf{F}^{S}_{L,a}, \mathbf{F}^{T}_{L,a})
\end{equation}
where $\mathbf{F}^{S}_{L,a}\in \mathbb{R}^{|x|\times d_{i}}$ and $\mathbf{F}^{T}_{L,a}\in \mathbb{R}^{|x|\times d_{i}}$ are mapping matrices of the the dimension increase and dimension reduction in feed forward module for student's and teacher's layer.
$d_{i}$ is the feed-forward size of model.

To comprehensively utilize the in-layer matrix information from both the self-attention and fully connected modules, we compute the overall matrix-aspect training loss by summing the attention matrix transfer loss and the mapping matrix transfer loss:
\begin{equation}
 \mathcal{L}_{\rm matrix}=\mathcal{L}_{\rm MHA}+\mathcal{L}_{\rm FNN}
\end{equation}

\begin{table*}[t]
\centering
\small
\fontsize{9.5pt}{12pt}\selectfont % 调整表格字体与行距
\begin{tabularx}{\linewidth}{lc|cccccccc|c}
\toprule
\textbf{Model}      & \textbf{\#Param} 
& \textbf{RTE}      &\textbf{MRPC}      &\textbf{STS-B}     &\textbf{CoLA} 
& \textbf{SST-2}    & \textbf{QNLI}     & \textbf{QQP}      & \textbf{MNLI-m/-mm}  
& \textbf{Avg}\\
\midrule
BERT$_{\text{BASE}}$ & 109M & 63.4 & 90.4 & 87.4 & 59.6 & 92.5 & 90.7 & 88.2 & 83.3/83.6 & 82.1  \\
\midrule
BERT$_{\text{SMALL}}$ & 29.2M & 58.3 & 83.1 & 82.0 & 29.5 & 89.9 & 82.2 & 83.4 & 77.6/77.2 & 73.6\\
% $\rm BERT_{4}\text{-}PKD$ & 52.2M & & & & & & & & \\
DistilBERT$_{4}$ & 52.2M & 55.7 & 82.6 & 82.9 & 31.4 & 90.2 & 82.7 & 83.8 & 78.9/78.1 & 74.0\\
TinyBERT$_{4}$ & 14.3M & 67.1 & 86.8 & 86.6 & 34.9 & 90.7 & 84.9 & 85.2 & 81.6/81.5 & 77.6\\
\textsc{MiniLM}v2 & 22.7M & 65.7 & 89.5 & 85.8 & 38.9 & 91.0 & 88.4 & 86.1 & 82.2/82.0 & 78.8\\
\text{MaKD}$_{64}$ & 24.1M & 66.4 & 89.6 & 85.5 & 40.9 & 91.0 & 88.9 & 86.7 & 82.4/82.3 & 79.3\\
\midrule
% $\rm BERT_{6}\text{-}PKD$ & 67.0M & 63.2 & 89.5 & 84.1 & 4& & & & \\
DistilBERT$_{6}$ & 67.0M & 63.7	& 90.2	& 86.9	& 51.3	& 90.7	& 87.2	& 87.2	& 82.0/82.2 & 80.1    \\
TinyBERT$_{6}$  & 67.0M  & \textbf{72.2} & 90.6 & \textbf{87.4} & 48.2 & 92.1  & 89.6 & 87.4 & \textbf{83.0}/82.7 & 81.4\\
\textsc{MiniLM}v2  & 67.0M & 67.5 & 91.2 & 87.3 & 48.6 & 92.2 & 89.9 & 87.6 & 82.8/82.6 & 81.0 \\
\textbf{MaKD}$_{\textbf{256}}$   & 67.0M & 68.2 & \textbf{91.3} & \textbf{87.4} & \textbf{55.9} & \textbf{92.5} & \textbf{90.2} & \textbf{87.8} & \textbf{83.0/82.8} & \textbf{82.0}\\
\bottomrule
\end{tabularx}
\caption{Results of our student on development sets of GLUE. We report F1 scores for MRPC, Person correlation for STS-B, Matthew's correlation for CoLA and accuracy for other datasets. 
The GLUE results of DistilBERT are taken from \cite{sanh2020distilbertdistilledversionbert}. For TinyBERT and \textsc{MiniLM}v2, we fine-tune the lastest version of their public model for a fair comparison. 
The results of our fine-tuning experiments are an average of 5 runs for each task. The best results are in \textbf{bold}.}
\label{tab:main}
\end{table*}

\paragraph{Model Distillation}
In addition to imitating the behaviors of intermediate layers, we also use the knowledge distillation to fit the predictions of teacher model as in \cite{hinton2015distillingknowledgeneuralnetwork}. Specifically, we perform soft - label distillation on the logit prediction layers of the two pre - training tasks, namely the Masked Language Model (MLM) and the Next Sentence Prediction (NSP).

\begin{equation}
    \mathcal{L}_{\rm model}=CE(\mathbf{z}^{T}/t,~\mathbf{z}^{S}/t)
\end{equation}
where $\mathbf{z}^{S}, \mathbf{z}^{T} \in \mathbb{R}^{C}$ are the logits vectors predicted by the student and teacher respectively, $C$ represents the number of classes. $CE()$ means the cross entropy loss, and $t$ means the temperature value.

\subsection{Multi-Aspect Learning}
% 我们在浅层网络使用矩阵蒸馏学习知识的表层信息，在深层网络通过层蒸馏表证句法特征。
We use matrix distillation to learn the surface information of knowledge in shallow networks, and use layer distillation to demonstrate syntactic features in deep networks.
Using the above distillation objectives (i.e. Equations 14, 17, 18), we can unify the distillation loss of the corresponding layers between the teacher and the student network:
\begin{equation}
\rm \mathcal{L}=\left\{
\begin{aligned}
& \rm \mathcal{L}_{matrix}\text{,}  & 0 \textless m \leq \tfrac{L}{2}\\
& \rm \mathcal{L}_{layer} \text{,}   & \tfrac{L}{2}\textless m \leq L \\
& \rm \mathcal{L}_{model}\text{,}   & m=L+1 \\
\end{aligned}
\right.
\end{equation}

where $m$ represent the layer of distillation. $L$ is the number of Transformer layer.

\begin{table}[t]
\centering
\small
\fontsize{10.4pt}{13pt}\selectfont 
\begin{tabularx}{\linewidth}{l|ccccc}
\toprule
\textbf{Model}      &\textbf{SQuAD 1.1} & \textbf{SQuAD 2.0}    & \textbf{Avg} \\
\midrule

BERT$_{\text{BASE}}$   & 76.0/83.0  &  69.5/72.6 &  75.2 \\
\midrule

DistilBERT$_6$
& 67.7/76.7  & 62.9/64.3  &  67.9  \\

TinyBERT$_6$
& 68.2/77.4  &  65.8/69.6 &  70.2 \\

\textsc{MiniLM}v2
& 73.9/81.4 & 68.2/71.8 & 73.8 \\

\textbf{MaKD$_{\textbf{256}}$}
& \textbf{74.9/82.3} & \textbf{69.1/72.9} & \textbf{74.8} \\

\bottomrule
\end{tabularx}
\caption{Comparative Studies on SQuAD 1.1 and SQuAD 2.0. We report the exact match and F1 scores on the development set. The best results are in \textbf{bold}.}
\label{tab:squad}
\end{table}

\begin{table*}[htb]
% \small
% \fontsize{10pt}{12pt}\selectfont
\centering
\begin{tabular}{l|cccccccc}
\toprule
\textbf{Model}   &\textbf{SST-2}  & \textbf{MNLI-m}  
&\textbf{CoLA}   & \textbf{STS-B}   & \textbf{MRPC} & \textbf{QNLI} & \textbf{SQuAD 2.0} & \textbf{Avg} \\
\midrule
MaKD$_{256}$ & 92.5	&83.0	&55.9	&87.4	& 91.3 & 90.2 & 72.9 & 81.8\\
\midrule
$\text{w/o}~\mathcal{L}_{\rm matrix}$ &91.6	&81.9	&51.8	&86.5	& 89.6 & 88.7 & 71.5 &80.2\\
~~~~~~$\text{w/o}~\mathcal{L}_{\rm MHA}$ & 92.0 & 82.4 & 52.3 & 86.9 & 90.2 & 89.3 & 72.1 & 80.7\\
~~~~~~$\text{w/o}~\mathcal{L}_{\rm FNN}$ & 92.4 & 82.7 & 54.0 & 87.3 & 90.8 & 89.8 & 72.4 & 81.0\\
$\text{w/o}~\mathcal{L}_{\rm layer}$ & 92.2	&82.3	&53.3	&86.6	& 89.9 & 89.5 & 72.0 & 80.8\\
$\text{w/o}~\mathcal{L}_{\rm model}$ &90.9	&82.0	&49.2	&86.7	& 89.4 & 89.1 & 71.2 & 79.7\\
\bottomrule
\end{tabular}
\caption{Ablation studies of different distillation objects in the MaKD learning. The variants are validated on the development set of GLUE benchmark and SQuAD 2.0.}
\label{tab:ablation}
\end{table*}

\begin{table*}[t]
\centering
% \small
% \fontsize{7.8pt}{10pt}\selectfont
\begin{tabular}{@{\extracolsep{1em}}ccc|cccc}
% $\boldsymbol{L_{MD}}$ & $\boldsymbol{L_{LD}}$ & $\boldsymbol{L_{KD}}$ &\textbf{MRPC}  & \textbf{CoLA}    & \textbf{Avg} \\
\toprule
$\rm \mathcal{L}_{matrix}$ & $\rm \mathcal{L}_{layer}$ & $\rm \mathcal{L}_{model}$ &\textbf{MRPC}  & \textbf{STS-B} & \textbf{MNLI-m}  & \textbf{Avg} \\
\midrule
% - & - & - & 87.6	\\
$[1\text{,}~\frac{L}{2}]$ & - & - & \textbf{90.0} & \textbf{87.1} & \textbf{81.9} & \textbf{86.3}\\
$[\frac{L}{2}\text{,}~L]$ & - & - & 87.2 & 85.5 & 81.5 & 84.7\\
$[1\text{,}~L]$ & - & - & 89.4 & 86.9 & 81.5 & 85.9 \\
\midrule
- & $[1\text{,}~\frac{L}{2}]$ &  - & 87.5 & 85.5 & 81.5 & 84.8 \\
- & $[\frac{L}{2}\text{,}~L]$ & - & \textbf{89.9} & \textbf{86.8} & \textbf{82.0} & \textbf{86.2} \\
- & $[1\text{,}~L]$ & - & 89.7 & 86.6 & 81.7 & 86.0 \\
\midrule
$[1\text{,}~L]$ & $[1\text{,}~L]$ & $L$ & 90.7 & 86.7 & 81.9 & 86.4 \\
$[1\text{,}~\frac{L}{2}]$ & $[\frac{L}{2}\text{,}~L]$ & $L$ & \textbf{90.9} & \textbf{87.2} & \textbf{82.4} & \textbf{86.8} \\

\bottomrule
\end{tabular}
\caption{Experimental results of distillation objects acting on different layers. Here, $L$ denotes the number of layers of the Transformer-based model (i.e. $L$=12, when model is BERT$_{\text{BASE}}$ in this experiment).}
\label{tab:dif}
\end{table*}
% 蒸馏目标作用在不同层的实验结果。其中L代表Transformer模型的层数（在实验中使用的是BERT-base，L=12）

\section{Experiments}
In this section, we conduct distillation experiments on different architectural teacher models including BERT$\rm_{\text{BASE}}$ \cite{devlin2019bertpretrainingdeepbidirectional}, GPT-2$\rm_{\text{BASE}}$ \cite{radford2019language} and LLaMA-3 \cite{grattafiori2024llama3herdmodels}.
\subsection{Setup}
We take a pre-trained language model BERT$\rm_{\text{BASE}}$ \cite{devlin-etal-2019-bert} with 109M parameters as the teacher model (number of layers $L=12$, hidden size $d^{T}=768$, intermediate size $d_{f}^{T}=3072$, and attention head number $A_{h}^{T}=12$),  
For the pre-training data, we use English Wikipedia and BooksCorpus \cite{7410368}. We train student models using 512 as the batch size and 1e-4 as the peak learning rate for 400,000 steps. We use AdamW with $\beta_{1}=0.9, \beta_{2}=0.99$.
We conduct distillation experiments using 8 NVIDIA Tesla A100 GPUs with mixed precision training.
In the fine-tuning stage, we perform a grid search over the learning rates $\{1.0e^{-5}, 2.3e^{-5}, 3.5e^{-5}, 4.8e^{-5}\}$ and 5 different seeds and choose the best model according to the validation set of each task.

We selected several mainstream knowledge distillation methods as baselines:
\begin{itemize}
    \item \textbf{DistilBERT} \cite{sanh2020distilbertdistilledversionbert}, which distills the student by the combination of the original MLM loss, the cosine distance for features, and the KL divergence for output logits.
    \item \textbf{TinyBERT} \cite{jiao-etal-2020-tinybert}, which aligns the attention distributions and hidden states for general distillation. 
    \item \textbf{\textsc{MiniLM}v2} \cite{wang-etal-2021-minilmv2}, which align the attention matrix and values-values scaled dot-product.
We refer to our approach as MaKD to signify that it is Multi-aspect Knowledge Distillation.
\end{itemize}

\begin{table*}[t]
\centering
% \small
% \fontsize{8.2pt}{10pt}\selectfont 
\begin{tabular}{@{\extracolsep{1em}}cccc|cccc}
\toprule
\textbf{Model} & \textbf{Role} & \textbf{\#Param} & \textbf{Method}  & \textbf{Dolly} & \textbf{SelfInst} &\textbf{Vicuna} & \textbf{Avg} \\
\midrule
\multirow{6}*{\makecell{GPT-2}} & Teacher & 124M & SFT & 22.5 & 9.05 & 13.99 & 15.18 \\
\cmidrule(r){2-8}
~ & \multirow{5}*{\makecell{Student}} & \multirow{5}*{\makecell{77.4M}} & None & 5.69 & 4.19 & 9.51 & 6.46\\
~ & ~ & ~ & SFT & 15.47 & 6.31 & 12.24 & 11.34\\
~ & ~ & ~ & KD  & 16.25 & 6.18 & 11.82 & 11.41\\
~ & ~ & ~ &  SeqKD & 16.56  & 6.33  & 12.14 & 11.67\\
~ & ~ & ~ &  \textbf{MaKD} &\textbf{18.56} &\textbf{8.26} &\textbf{13.77} & \textbf{13.53}\\
\midrule[0.8pt]
\multirow{6}*{\makecell{LLaMA-3}}  & Teacher & 1.23B & SFT  & 27.47 & 18.80 & 18.01 & 21.42\\
\cmidrule(r){2-8}
~ & \multirow{5}*{\makecell{Student}} & \multirow{5}*{\makecell{0.69B}} & None & 0.50 & 0.55 & 0.96 & 0.67\\
~ & ~ & ~ &  SFT & 16.10 & 7.82 &  11.38 & 11.76\\
~ & ~ & ~ &  KD & 16.04 & 7.23 & 12.62 & 11.96 \\
~ & ~ & ~ &  SeqKD & 17.05 & 6.76 & 12.54 & 12.11\\
~ & ~ & ~ &  \textbf{MaKD} & \textbf{20.74}  & \textbf{9.38}  & \textbf{15.62} & \textbf{15.24}\\
\bottomrule
\end{tabular}
\caption{Experimental results on auto-regressive models (i.e., GPT-2 and LLaMA-3). We report Rouge-L score for three instruction following tasks. KD \cite{song2020lightpafftwostagedistillationframework} fine-tunes the student model on dataset using the teacher distribution as the supervision at each token step. SeqKD \cite{NEURIPS2023_ac662d74} fine-tunes the student model on the data generated by the teacher model.}
\label{tab:llama}
\end{table*}

\subsection{Downstream Tasks}
We evaluate our proposed approach on the Natural Language Understanding (NLU) and Natural Language Generation (NLG) datasets, respectively.
\paragraph{GLUE}
General Language Understanding Evaluation (GLUE) benchmark \cite{wang-etal-2018-glue} consists of 2 single sentence tasks: CoLA \cite{warstadt-etal-2019-neural}, SST2 \cite{socher-etal-2013-recursive}, 3 sentence similarity tasks: MRPC \cite{dolan-brockett-2005-automatically}, STS-B \cite{cer-etal-2017-semeval}, QQP, and 4 natural language inference tasks: MNLI \cite{williams-etal-2018-broad}, QNLI \cite{rajpurkar-etal-2016-squad}, RTE \cite{bentivogli2009fifth} and WNLI (Levesque et al., 2012).

\paragraph{Extractive Question Answering}
The task aims to predict a continuous sub-span of the passage to answer the question. SQuAD is a question answering dataset that include SQuAD 1.1 \cite{rajpurkar-etal-2016-squad} and SQuAD 2.0 \cite{rajpurkar-etal-2018-know}.

\paragraph{Instruction Following}
We evaluate our Muiti-KD framework on several instruction-following dataset following \cite{gu2024minillm}. Specifically, we choose databricksdolly-15k dataset processed by \cite{gu2024minillm} to conduct the KD process, which contains about 11k samples for training, 1k for validation, and 500 for testing. Besides, we also select Self-Instruct \cite{wang-etal-2023-self-instruct}, Vicuna-Evaluation \cite{peng2023instructiontuninggpt4} as the additional test sets for more comprehensive evaluation.

\subsection{Main Results}
As shown in Table~\ref{tab:main}, MaKD achieves competitive performance, which is distilled from BERT$_\text{BASE}$ on GLUE development sets.
In particular, MaKD retains 99.8\% and 96.5\% performance of BERT$_\text{BASE}$ using only 61.4\% and 22.1\% parameters, respectively.
Compared to the baselines with 67.0M parameters, MaKD$_{256}$ get higher performance than TinyBERT and \textsc{MiniLM}v2, showing its superiority over the traditional KD methods only with logit-based loss and feature-based loss.
In particular, our model exceeds the two stage-of-the-art models by 7.3\% accuracy on CoLA.
However, MaKD's performance on the RTE task slightly lags behind TinyBERT, likely due to the relatively small scale of the RTE dataset, which may not provide sufficient diverse samples for MaKD to fully capture the semantic relationships and inference patterns it is designed to learn. 
We also conduct experiments for smaller student models. MaKD$_{64}$ significantly outperforms the 4-layer state-of-the-art KD baselines with only 24.1M parameters. 

The comparative results evaluated on the development sets of SQuAD 1.1 and 2.0 are presented in Table \ref{tab:squad}, with exact match and F1 scores for evaluation. The results show that MaKD outperformes all baseline methods under both parameter settings, which further demonstrates the effectiveness of our method.

\subsection{Ablation Studies}
% 我们进行消融研究来分析不同方面对KD的贡献。四个任务的开发结果如表x所示。矩阵蒸馏（MatD）、层蒸馏（LayD）和模型蒸馏（ModD）对最终结果有积极的贡献。提取矩阵方面的细粒度知识有助于学生模型深度模仿教师的自注意力行为，从而进一步提高模型的性能。
As shown in Table~\ref{tab:ablation}, we perform ablation studies to analyze the contribution of different aspects in KD. The matrix distillation ($\mathcal{L}_{\rm matrix}$), layer distillation ($\mathcal{L}_{\rm layer}$), and model distillation ($\mathcal{L}_{\rm model}$) positively contribute to the final results. Specifically, the average performance $\text{w/o}~\mathcal{L}_{\rm matrix}$ drops significantly from 81.8\% to 80.2\%. Distilling the fine-grained knowledge of matrix aspect helps the student model deeply mimic the self-attention behavior of the teacher, which further improves model performance.
% 值得注意的是，the variants w/o model 相比于MaKD256在平均结果上下降了2.1%, 这说明对齐带有高级语义的logits信息的也是至关重要的。
It is worth noting that the variants w/o model shows a 2.1\% decrease in the average results compared to MaKD$_{256}$. This indicates that aligning the logits information with high - level semantics is also of crucial importance.
Furthermore, we delve into the contributions made by the linear layers within the Multi-Head Attention ($\mathcal{L}_{\rm MHA}$) and Feed-Forward Neural Network ($\mathcal{L}_{\rm FNN}$) during the matrix distillation process. Our findings reveal that the distillation based on MHA exerts a more substantial influence compared to that based on FNN. Meanwhile, these two types of knowledge distillation complement one another. As a result, in our experiments, they emerge as the most crucial distillation techniques for Transformer-based models.

\section{Analysis and Discussion}

\subsection{Effect of different layers}
% 我们在三个任务上分别探究了对不同层进行矩阵蒸馏和层蒸馏的效果。正如表4所示，在矩阵蒸馏时选择浅层网络的平均结果为86.3，这要高于对深层网络和全部层进行蒸馏的结果84.7和85.9。与此同时，在层间蒸馏时选择深层网络的平均结果为86.2，比选择浅层网络的结果要高出1.4%.
% 此外，我们发现当在浅层网络和深层网络分别进行矩阵蒸馏和层蒸馏，并添加模型蒸馏的时候，可以取得最优的结果。这证明了通过在浅层网络学习局部特征并在深层网络传递高层语义知识的分层蒸馏方式有利于捕捉到更丰富的教师行为信息。
We explored the effects of matrix distillation ($\mathcal{L}_{\rm matrix}$) and layer distillation ($\mathcal{L}_{\rm layer}$) on different layers respectively across three tasks. As shown in Table~\ref{tab:dif}, when performing matrix distillation, the average result of selecting the shallow-layer network is 86.3\%, which is higher than the results of 84.7\% and 85.9\% obtained from distilling the deep-layer network and all layers. Meanwhile, when performing layer distillation, the average result of choosing the deep-layer network is 86.2\%, which is 1.4\% higher than that of choosing the shallow-layer network.
In addition, we found that when matrix distillation and layer distillation are carried out on the shallow-layer network and the deep-layer network respectively, and model distillation is added, the optimal result can be achieved. This demonstrates that the hierarchical distillation approach, which involves learning local features in the shallow-layer network and transmitting high-level semantic knowledge in the deep-layer network, is conducive to capturing more abundant teacher behavior information.

\subsection{Results on Auto-Regressive}
% 图\ref{tab:llama}给出了不同自回归模型（即GPT2-124M和LLaMA3-1B）提取的MacKD结果。特别地，我们使用Rouge-L 分数来衡量模型生成的精度。
Figure \ref{tab:llama} presents the results of MaKD distilled from different auto-regressive models (i.e. GPT-2 and LLaMA-3). We use the Rouge-L \cite{lin-2004-rouge} score to measure the precision of the model generation. 
% 特别地，我们分别采用监督微调的GPT2-124M和LLaMA3-1B作为与学生llm具有相同词汇的教师llm。对于学生llm，我们使用低秩分解来初始化教师模型。
Specially, we adopt supervised fine-tuned GPT2-124M and LLaMA3-1B respectively as the teacher LLMs that have the same vocabularies with the student LLMs. For student LLMs, we init the teacher models with low-rank factorization. 
% 值得一提的是MAKD在自回归模型上展示了出色的性能优势。对于GPT2-124M，学生模型以62%的参数量实现了教师模型82%的性能。在同等参数设置的情况下，MaKD的表现优于其他极限方法，如Supervise fine tuning（SFT）、Knowledge Distillation（KD）和SeqKD。
It is worth mentioning that MaKD demonstrates excellent performance advantages on auto-regressive models. For GPT2-124M, the student model achieves 82\% of the performance of the teacher model with only 62\% of the parameters. Under the same parameter settings, MaKD outperforms other baselines, such as Supervise fine tuning (SFT), Knowledge Distillation (KD) and SeqKD.
% 对于LLaMA-3的参数量从1.23B减少为0.69B，MaKD在三个指令遵循数据集上均取得了最佳结果。实现了。学生模型初始化后
When the number of parameters of LLaMA-3 is reduced from 1.23B to 0.69B, MaKD achieves the best results on all three instruction-following datasets. 

% 我们的结果证明了在自回归模型上的适用性，这为LLMs模型压缩的后续研究提供了坚实的理论支撑。
Our results demonstrate the applicability of MaKD on auto-regressive models, furnishing a solid theoretical underpinning for the subsequent research on the compression of large language models (LLMs).

\section{Related Work}
Knowledge Distillation \cite{hinton2015distillingknowledgeneuralnetwork} is a widely used technique to transfer knowledge from a large model (teacher) to a smaller one (student) to achieve all types of efficiency. Based on the loss function designs, we can further categorize the methods into logit-based KD and feature-based KD. 

\paragraph{Logit-based KD} Logit-based KD uses the KL divergence or mean squared error (MSE) to minimize the logits between the student and the teacher. \cite{tang2019distillingtaskspecificknowledgebert} distills fine-tuned BERT into a BiLSTM model in a task-specific setting. DistilBERT \cite{sanh2020distilbertdistilledversionbert} uses soft target probabilities and embedding outputs to train student model. SeqKD \cite{kim-rush-2016-sequence,NEURIPS2023_ac662d74} fine-tunes the student model on the sequence-level teacher-generated data. MixKD \cite{liang2021mixkdefficientdistillationlargescale} extends the idea of encouraging the student to mimic the teacher’s logits to the linear interpolation of example pairs \cite{zhang2018mixupempiricalriskminimization}. PD \cite{turc2019wellreadstudentslearnbetter} find that a student BERT pretrained with MLM outperforms random initialization and truncated teacher when used to initialize the student model.

\paragraph{Feature-based KD }Instead of using only the final output, feature-based KD aims to align the intermediate features between the teacher and the student. BERT-PKD \cite{sun-etal-2019-patient} learns from multiple intermediate layers of teacher model for incremental knowledge extraction. TinyBERT \cite{jiao-etal-2020-tinybert} and MobileBERT \cite{sun-etal-2020-mobilebert} further utilize self-attention distributions and outputs of each Transformer layer. \cite{NEURIPS2020_6f5216f8} uses layerwise KD loss to distill a teacher into a student model that has sub-networks of different widths and depths. \textsc{MiniLM} \cite{NEURIPS2020_3f5ee243} use self-attention distribution and value relation to conduct deep self-attention distillation. \textsc{MiniLM}v2 \cite{wang-etal-2021-minilmv2} generalizes deep self-attention distillation in \textsc{MiniLM}, employing self-attention relation.

\section{Conclusion}
In this research, we investigate multi-aspect knowledge distillation on the Transformer-based models. We introduce MaKD through a hierarchical distillation strategy, a novel approach improves knowledge alignment during the distillation process by capturing detailed aspects of the internal representations of the model.
Extensive experimental verification shows that our method is applicable to various Transformers-based models,  including both encoder-only and auto-regressive decoder-only architectures with different sizes. 
In the future, we would examine its applicability in a broader range of domains and scenarios.

\section*{Limitations}
% 限制1：引入了矩阵蒸馏后训练成本更高了
% 限制2：随着scaling law，LLMs模型越来越大，我们尚未探究在更高压缩率（10%）下的表现。
% 限制3: 此外，我们实验使用的LLaMA3是最小规模的1B，没有构造在7B/13B上更大LLMs的实验。
One limitation is that, the multi-aspect distillation that combines coarse and fine granularities requires more training time than previous methods. 
Furthermore, the models used in our experiments were limited in scale, as we have not conducted experiments on larger-scale LLMs nor explore the performance under a lower compression rate (<10\%). To address this limitation, future work could explore the potential of MaKD in these complex NLP domains.

\bibliography{custom}

\appendix

% \section{Example Appendix}
% \label{sec:appendix}

% This is an appendix.

\end{document}